\documentclass{article}
\usepackage[T1]{fontenc}
\usepackage{icomp2026_conference,times}

\usepackage{amsmath,amsfonts,bm}

\def\eqref#1{equation~\ref{#1}}

\def\1{\bm{1}}

\DeclareMathAlphabet{\mathsfit}{\encodingdefault}{\sfdefault}{m}{sl}
\SetMathAlphabet{\mathsfit}{bold}{\encodingdefault}{\sfdefault}{bx}{n}

\newcommand{\E}{\mathbb{E}}

\newcommand{\R}{\mathbb{R}}

\usepackage{hyperref}
\usepackage{url}
\usepackage{graphicx}
\usepackage{booktabs}
\usepackage{amsmath,amssymb,amsthm}
\usepackage{array}
\usepackage{algorithm}
\usepackage{algpseudocode}
\usepackage{float}
\usepackage{placeins}
\hypersetup{hidelinks,hypertexnames=false,
  pdftitle={Policy-Mask: Budget-Aware Adaptation of Diffusion Features to Discriminative Tasks},
  pdfauthor={Ivan Listopadov, Maksim Nakhodnov, Aibek Alanov}}

\newcommand{\Policy}{\textsc{Policy-Mask}}
\newcommand{\given}{\,|\,}

\newcommand{\IvanAffiliation}{1}
\newcommand{\MaksimAffiliation}{2, 3, 4}
\newcommand{\AibekAffiliation}{1, 4}
\makeatletter
\newcommand{\algbottomskip}[1]{\gdef\@fs@post{\kern2pt\hrule\relax\vskip#1}}
\makeatother

\title{Policy-Mask:\ Budget-Aware Adaptation of Diffusion Features to Discriminative Tasks}

\author{Ivan Listopadov\textsuperscript{\IvanAffiliation}
\And
Maksim Nakhodnov\textsuperscript{\MaksimAffiliation}
\And
Aibek Alanov\textsuperscript{\AibekAffiliation}
\AND
\textnormal{\textsuperscript{1}HSE University}\and 
\textsuperscript{2}MSU\and
\textsuperscript{3}Constructor University\and
\textsuperscript{4}FusionBrain Lab, AXXX
}

\icompfinalcopy

\begin{document}
\maketitle

\fancyhead{}
\renewcommand{\headrulewidth}{0pt}

\begin{abstract}
Selecting intermediate features from a frozen diffusion model for
discriminative tasks is non-trivial because the useful signal is distributed
across the denoising trajectory, network depth, and channels. In practice, this
requires choosing a timestep, a block, and a channel subset. Existing methods
either fix these choices by hand or optimize only a few extraction
hyperparameters, most often the timestep, then reuse the same rule across
inputs and budgets. We introduce \Policy{}, a budget-conditioned policy
trained on the downstream objective that jointly selects all three under a
prescribed channel budget. Without normalization, the channel-score
contribution grows with the number of selected channels and dominates the
timestep and block updates. We normalize it by subset cardinality, keeping the
three policy heads on comparable scales and stabilizing joint training. At
matched budgets, \Policy{} improves over fixed extraction and feature-filter
baselines across segmentation, classification, and semantic correspondence,
reaching 38.9 mIoU on ADE20K, 75.4\% top-1 on CIFAR-100 at 128 dimensions, and
76.3 PCK@0.1 on SPair-71k. Component-wise ablations show that block selection
materially affects downstream accuracy, and the highest recorded channel-budget
score occurs at 256 retained channels, identifying a practical operating point
for compact diffusion representations. We further adapt \Policy{} to
diffusion transformers, including PixArt-$\Sigma$ and SD3.5 MMDiT, by
replacing U-Net channels with image-token hidden dimensions.
\end{abstract}

\section{Introduction}

Diffusion models learn intermediate representations that support semantic
segmentation, image classification, and semantic correspondence
\citep{baranchuk2022,xiang2023,tang2023}. Using these representations for a
downstream task requires choosing which activations to extract. The available
features depend on the denoising timestep and network layer, and each feature
map contains a set of channels from which the task representation is formed.

Existing methods address this choice through fixed extraction rules, learned
aggregation across timesteps and layers, or timestep selection during
distillation \citep{luo2023,yang2023}. RepFusion transfers features from a
frozen diffusion teacher to a student and learns which timestep to use. This
leaves the choice of layer and retained feature dimensions as further parts
of the extraction problem. These choices also determine which activations
must be stored and used as distillation targets.

\Policy{} selects a timestep, a layer, and a channel subset under a
prescribed channel budget. The policy takes the input image and budget and
selects an action $a=(t,l,K)$ with $|K|=b$. A task decoder provides the
selection reward, and the selected features supervise the student during
training. At inference, the student processes the image without evaluating
the diffusion teacher. This extends RepFusion's timestep selection to a joint
choice of extraction point and feature subset.

Our contributions are:
\begin{enumerate}
\item A budgeted $(t,l,K)$ policy with exact-cardinality sampling.
\item A channel-score normalization that controls the relative action-head contributions.
\item At 128 of 1280 aligned channels, gains of 1.2--1.6 mIoU over full-width
fixed extraction on four segmentation benchmarks.
\item A formal action mapping from U-Net features to DiT, MMDiT, and flow
transformer feature lattices.
\end{enumerate}

Section~\ref{sec:related} reviews related work, Section~\ref{sec:method}
defines the policy, Section~\ref{sec:experiments} reports experiments, and
Section~\ref{sec:conclusion} concludes. Appendices~\ref{app:details}--\ref{app:qualitative}
provide experimental details and additional results.

\section{Related work}
\label{sec:related}

\textbf{Diffusion representations.}
Diffusion features support dense prediction, correspondence, segmentation, and
depth estimation through task-specific extraction rules
\citep{baranchuk2022,xiang2023,yang2023,luo2023,tang2023,zhao2023,ji2023,
xu2023odise,li2023dreamteacher,ke2024}. Recent work studies prompt selection,
noise-free representations, and the transfer of stable diffusion to dense
tasks \citep{li2024sd4match,tian2024diffseg,kondapaneni2024,zhang2024dense,
stractke2025,ravishankar2025}. RepFusion is the closest predecessor: it
distills a recognizer from a frozen diffusion teacher after selecting a useful
noise step \citep{yang2023}. \Policy{} keeps the same teacher--student
division and expands the selected object to $(t,l,K)$.

\textbf{Budgeted subset selection.}
Feature selection spans filters, wrappers, continuous relaxations, and greedy
methods \citep{peng2005,kohavi1997,abid2019,daskempe2011,elenberg2018,
mirzasoleiman2015,nemhauser1978}. \Policy{} selects feature dimensions of one frozen model response.
Budget-conditioned policies also appear in diffusion caching, where ReCache
selects recomputation steps under an inference budget \citep{aliev2026recache}.
Here the budget constrains feature coordinates used for distillation.
The policy uses a Plackett--Luce ordering and a Gumbel top-$b$ sampler
\citep{plackett1975,kool2019}. Its gradient estimator follows the standard
score-function construction with a baseline \citep{greensmith2004,kool2020}.

\section{Method}
\label{sec:method}

\Policy{} learns which teacher response to distill under an exact channel
budget. The task decoder evaluates the selected response, and its loss supplies
the policy reward while the selected feature supervises the student.

\subsection{Problem setup}

Let $\mathcal{D}$ be a data distribution over $(x,y)$ and let the frozen
teacher $D$ expose $F_{t,l}(x)\in\R^{C_{t,l}\times H_{t,l}\times W_{t,l}}$.
An alignment map $A_l$ projects a candidate layer to a common feature space,
$G_{t,l}(x)=A_l(F_{t,l}(x))\in\R^{C\times H_s\times W_s}$. This makes a
single exact budget meaningful even when raw layers have different widths and
resolutions. In the ADE20K experiments with trainable $A_l$, the KD loss fell below
0.005 within two epochs. We therefore fixed $A_l$ at initialization for the
normalization ablation and the RepFusion comparison.
Appendix~\ref{app:alignment} provides the experimental details. During training, $D$ is queried, or its responses are drawn
from a feature bank. At inference, the student alone consumes $x$
\citep{yang2023,huang2023}.
For budget $b$, the feasible action set is:
\begin{equation}
\mathcal A_b=\{(t,l,K):t\in\mathcal T,\ l\in\mathcal L,\
K\subseteq[C],\ |K|=b\}.
\label{eq:action}
\end{equation}
With a binary channel mask $m_K$, the selected response is
$\widetilde F_a(x)=G_{t,l}(x)\odot m_K$. The student returns task logits
$u_\phi(x)$ and a projected feature $p_\phi(x)$ and is trained with:
\begin{equation}
\mathcal L_{\rm stu}=\ell_{\rm task}(u_\phi(x),y)+
\lambda\ell_{\rm KD}\!\left(p_\phi(x),\operatorname{sg}[\widetilde F_a(x)]\right),
\label{eq:student-loss}
\end{equation}
where $\operatorname{sg}$ stops gradients through the frozen target. A small
task decoder $q_\omega$ maps the selected teacher feature to task predictions. Holding
$\omega$ fixed during a policy step, the selector objective is:
\begin{equation}
J_b(\theta)=\E_{(x,y)\sim\mathcal D,\ a\sim\pi_\theta(\cdot\given x,b)}
\left[-\ell_{\mathrm{dec}}\!\left(q_\omega(\widetilde F_a(x)),y\right)\right].
\label{eq:objective}
\end{equation}

\begin{figure}[t]
\centering
\includegraphics[width=\linewidth]{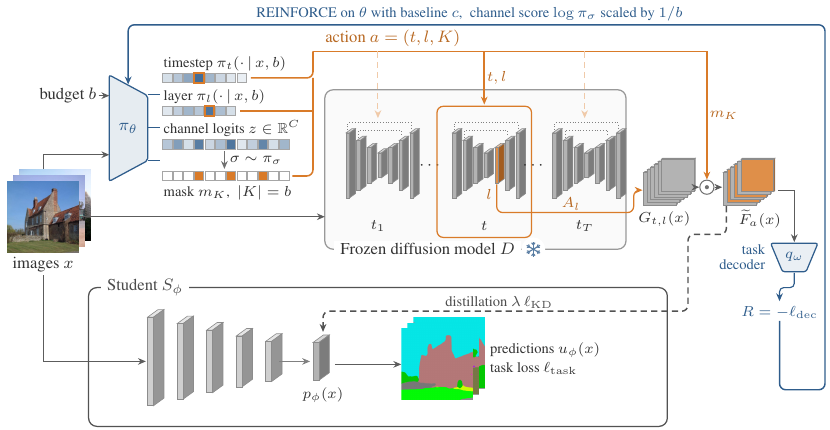}
\vspace{-1em}
\caption{\Policy{} training and inference. Given an image $x$ and a budget
$b$, the policy $\pi_\theta$ outputs distributions over timesteps and layers
and channel logits $z$. A timestep $t$ and a layer $l$ are sampled.
An ordered sequence $\sigma\sim\pi_\sigma$ of $b$ channels is drawn without
replacement by Gumbel top-$b$, giving the mask $m_K$. The frozen diffusion model $D$
returns $G_{t,l}(x)$ after the alignment $A_l$, and masking gives
$\widetilde F_a(x)$. This feature is the distillation target for the student
$S_\phi$ and, through the task decoder $q_\omega$, gives the reward
$R=-\ell_{\mathrm{dec}}$. The policy is updated by REINFORCE with baseline $c$
and the channel score scaled by $1/b$ (Equation~\ref{eq:balanced}). At inference, only
$S_\phi$ is evaluated. Images and predictions are ADE20K validation examples
(Figure~\ref{fig:qualitative}).}
\label{fig:overview}
\vspace{-1em}
\end{figure}

Both $\ell_{\rm task}$ and $\ell_{\mathrm{dec}}$ use the training annotation $y$:
pixel-wise supervision for segmentation, cross-entropy for classification, and
the corresponding matching loss. On ADE20K, $\ell_{\mathrm{dec}}$ averages
cross-entropy over valid pixels per image. The student task loss also includes
Dice loss.

The policy therefore receives one scalar reward
per sampled action, while held-out mIoU, top-1 accuracy, and PCK remain
evaluation metrics. The teacher is not differentiated through.
The action is discrete, and the feasible set has
$|\mathcal A_b|=|\mathcal T|\,|\mathcal L|\binom{C}{b}$ elements.
Rather than enumerating subsets, we sample actions and use a policy gradient.

\subsection{Policy over budgeted subsets}

The policy samples an ordered index sequence
$\sigma=(k_1,\ldots,k_b)\in\operatorname{Perm}_b([C])$ and then uses the
unordered set $K(\sigma)=\{k_1,\ldots,k_b\}$ to form the mask. It factorizes as:
\begin{equation}
\pi_\theta(t,l,\sigma\given x,b)=
\pi_t(t\given x,b)\,\pi_l(l\given x,b)\,\pi_\sigma(\sigma\given x,b).
\label{eq:factorization}
\end{equation}
The first two factors are categorical. For channel logits $z\in\R^C$, we draw
independent Gumbel perturbations and take the $b$ largest perturbed logits
in descending order:
\begin{equation}
g_c=-\log(-\log U_c),\quad U_c\overset{\rm iid}{\sim}\operatorname{Uniform}(0,1),
\qquad \sigma=\operatorname{Top}_b(z+g).
\label{eq:gumbel}
\end{equation}
This produces an ordered draw $(k_1,\ldots,k_b)$ without replacement. Its
exact Plackett--Luce score is:
\begin{equation}
\log\pi_\sigma(\sigma\given x,b)
=\sum_{j=1}^{b}\left[z_{k_j}-
\log\!\!\sum_{c\notin\{k_1,\ldots,k_{j-1}\}}\!\!\exp z_c\right].
\label{eq:pl}
\end{equation}
The reward is invariant to the order of $\sigma$, but the score is evaluated
on the draw that generated the mask. Top-$b$ therefore returns $b$ distinct
coordinates with exact feasibility rather than a sparsity penalty. Holding the
decoder fixed, an action-independent baseline $c$ gives the score-function
estimator \citep{greensmith2004,kool2020}:
\begin{equation}
\widehat g=(R-c)\nabla_\theta\log\pi_\theta(t,l,\sigma\given x,b),
\qquad \E[\widehat g]=\nabla_\theta J_b.
\label{eq:policy-gradient}
\end{equation}

\begin{algorithm}[t]
\algbottomskip{-1em}
\caption{\Policy{}: training and task inference}
\label{alg:policy}
\begin{algorithmic}[1]
\Require frozen teacher $D$, student $S_\phi$, task decoder $q_\omega$, policy $\pi_\theta$, budget $b$
\For{each minibatch $(x,y)$}
  \State sample $t,l,\sigma$ from $\pi_\theta(\cdot\mid x,b)$ and set $K\gets K(\sigma)$
  \State $\widetilde F \gets A_l(F_{t,l}^{D}(x))\odot m_K$
  \State $R \gets -\ell_{\mathrm{dec}}(q_\omega(\widetilde F),y)$
  \State update $\theta$ with $(R-c)\nabla_\theta[\log\pi_t+\log\pi_l+b^{-1}\log\pi_\sigma]$
  \State update $\phi$ with $\mathcal L_{\rm stu}$ and $\omega$ with $\ell_{\mathrm{dec}}(q_\omega(\operatorname{sg}[\widetilde F]),y)$
\EndFor
\State \textbf{Inference:} return $u_\phi(x)$ without evaluating $D$
\end{algorithmic}
\end{algorithm}

The implementation uses a stopped moving baseline for variance reduction. The task decoder supplies $R=-\ell_{\mathrm{dec}}$ and is trained on the same
decoder loss with its input detached. This separates selection from the student's
task and distillation losses. Emitting action logits costs
$O(|\mathcal T|+|\mathcal L|+C)$, and Gumbel perturbation plus top-$b$
selection is linear in $C$ up to the partial-sort cost. A frozen teacher query
and layer alignment dominate these selector operations.

\subsection{Channel-score normalization}

The channel score in Equation~\ref{eq:pl} is a sum of $b$ terms, whereas the
timestep and layer scores each contain one term. Dividing the channel score
by $b$ replaces this sum with its mean and controls its relative weight in
the joint update. It does not assume that gradient norms grow linearly with
$b$. We use:
\begin{equation}
\widehat g_{\rm bal}=(R-c)\nabla_\theta\left[
\log\pi_t+\log\pi_l+\frac{1}{b}\log\pi_\sigma\right].
\label{eq:balanced}
\end{equation}
The $1/b$ factor is a preconditioner, so the controlled update is not claimed
to be an unbiased estimator of $\nabla_\theta J_b$. For disjoint heads and
plain SGD, it is equivalent to reducing only the channel-head
learning rate by $b$. With a shared trunk, adaptive moments, and entropy
regularization, it instead changes the relative head contribution
\citep{gradnorm2018}. The action-axis ablation motivates this control: the
unnormalized joint policy reaches 31.68 mIoU, compared with 31.78 for the
timestep-only policy (Figure~\ref{fig:action-ablation}). With fixed alignment,
normalization raises mean best-validation mIoU from 36.8 to 37.7 over three
seeds (Table~\ref{tab:fixed-alignment}).

\section{Experiments}
\label{sec:experiments}

\subsection{Experimental setup}

\textbf{Teacher and student.}
The main selector comparisons use a frozen SD-1.5 U-Net and a common width
of 1280 aligned channels. Compact selectors retain 128 channels. The student
loss and updates are given in Equation~\ref{eq:student-loss} and
Algorithm~\ref{alg:policy}.

\textbf{Datasets and metrics.}
Segmentation is evaluated on ADE20K \citep{zhou2017}, Cityscapes
\citep{cordts2016}, CelebA-HQ \citep{karras2018pggan}, and Pascal VOC
\citep{everingham2010voc} using mIoU. Classification uses CIFAR-10 and
CIFAR-100 \citep{krizhevsky2009} top-1 accuracy. Semantic correspondence uses
SPair-71k \citep{min2019spair}
PCK@0.1 with image and bounding-box normalization reported separately.

\textbf{Baselines.}
The baselines include fixed extraction, top-variance and Fisher channel
selection, and full-width extraction. The DDAE \citep{xiang2023} and
RepFusion \citep{yang2023} entries in Table~\ref{tab:recognition} retain
their corresponding source protocols.

\subsection{Main results}

At 128 selected channels, \Policy{} exceeds full-width extraction at a fixed
$(t,l)$ by 1.5, 1.6, 1.2, and 1.3 mIoU on ADE20K, Cityscapes, CelebA-HQ,
and Pascal VOC, respectively (Table~\ref{tab:segmentation}). These results
compare feature-selection rules at 10\% of the full aligned width.

\begin{table}[t]
\caption{Segmentation results with different feature selection methods, measured
in mIoU. All methods use the same SD-1.5 feature lattice with $b=128$, except
for the ``All channels'' baseline, which retains all 1280 channels.
Bold values indicate the best result among 128-channel selectors.}
\vspace{1em}
\label{tab:segmentation}
\centering
\begin{tabular*}{\linewidth}{@{\extracolsep{\fill}}lrcccc@{}}
\toprule
Selector & Width & ADE20K & Cityscapes & CelebA-HQ & Pascal VOC \\
\midrule
All channels at fixed $(t,l)$ & 1280 & $37.4$ & $68.6$ & $77.9$ & $73.5$ \\
\midrule
Fixed $(t,l)$ + fixed $K$ & 128 & $33.8$ & $62.9$ & $72.4$ & $68.1$ \\
Top-variance $K$ & 128 & $35.6$ & $64.8$ & $74.6$ & $70.3$ \\
Fisher proxy $K$ & 128 & $36.4$ & $65.7$ & $75.3$ & $71.0$ \\
\Policy{} $(t,l,K)$ & 128 & $\mathbf{38.9}$ & $\mathbf{70.2}$ & $\mathbf{79.1}$ & $\mathbf{74.8}$ \\
\bottomrule
\end{tabular*}
\vspace{-1em}
\end{table}

\begin{table}[t]
\caption{Classification and correspondence comparison with DDAE
\citep{xiang2023} and RepFusion \citep{yang2023}. Full-width diffusion
methods and compact selectors are listed separately. A dash denotes a
method--task pair not reported under the corresponding source protocol.
Bold values indicate the best result among 128-dimensional selectors.}
\vspace{1em}
\label{tab:recognition}
\centering
\begin{tabular*}{\linewidth}{@{\extracolsep{\fill}}llccccc@{}}
\toprule
Method & Teacher & Width & CIFAR-10 & CIFAR-100 & \multicolumn{2}{c}{SPair-71k PCK@0.1} \\
\cmidrule(lr){6-7}
 & & & & & img & bbox \\
\midrule
DDAE & DDPM & full & $95.4$ & $76.8$ & --- & --- \\
RepFusion & SD-1.5 & full & $94.2$ & $74.1$ & --- & --- \\
\midrule
Top-variance $K$ & SD-1.5 & 128 & $93.1$ & $72.3$ & --- & --- \\
Fisher $K$ & SD-1.5 & 128 & $93.6$ & $73.1$ & --- & --- \\
\Policy{} & SD-1.5 & 128 & $\mathbf{95.2}$ & $\mathbf{75.4}$ & $\mathbf{76.3}$ & $\mathbf{67.5}$ \\
\bottomrule
\end{tabular*}
\end{table}

At 128 dimensions, \Policy{} reaches 95.2\% top-1 on CIFAR-10 and 75.4\%
on CIFAR-100, 1.0 and 1.3 points above the reported RepFusion entries
(Table~\ref{tab:recognition}). On SPair-71k, it reaches 76.3 PCK@0.1 with
image normalization and 67.5 with bounding-box normalization.

\begin{figure}[!t]
\centering
\includegraphics[width=\linewidth]{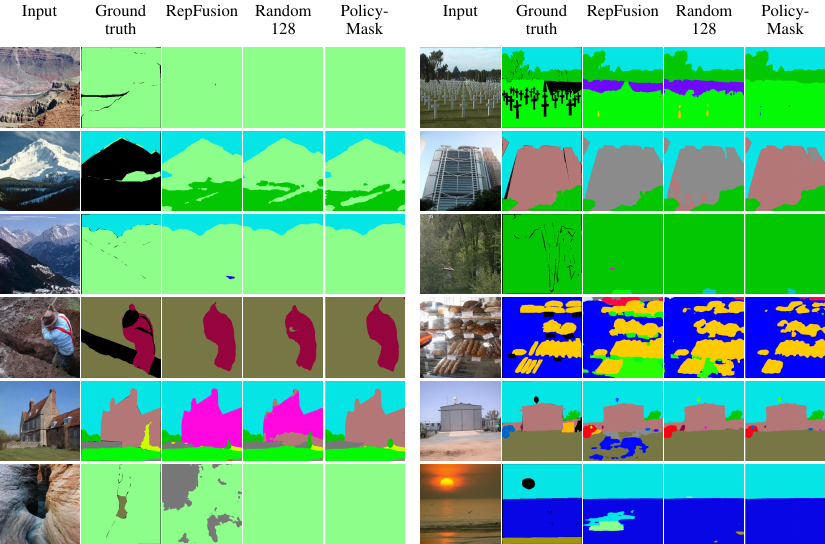}
\vspace{-1em}
\caption{ADE20K validation, seed 0. The 12 images with the largest gains
in image-mIoU over RepFusion. Selected by outcome. See the loss panel
(Figure~\ref{fig:panel-loss}). Columns in each block show input, ground truth,
RepFusion, Random 128, and \Policy{}.}
\label{fig:qualitative}
\vspace{-1em}
\end{figure}

Figure~\ref{fig:qualitative} shows corrected facade labels and fewer false
regions in cemetery, bakery, and sea scenes among the outcome-selected examples.
The corresponding losses and class-wise differences are reported in
Appendix~\ref{app:qualitative}.

\subsection{Ablation study}

\begin{figure}[!t]
\centering
\includegraphics{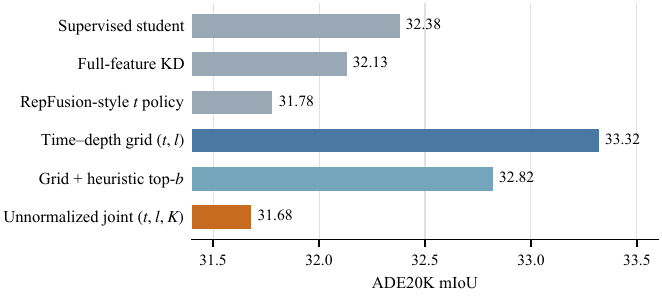}
\caption{Controlled action-axis ablation on ADE20K, mIoU. The time--depth
grid exceeds the timestep-only policy. Appending a heuristic subset does not
retain that gain, while the unnormalized joint action is lower. The values come
from one ablation run and are not compared numerically with Table~\ref{tab:segmentation}.}
\label{fig:action-ablation}
\end{figure}

\textbf{Action axes.}
In the controlled action-axis ablation, expanding a
timestep-only policy to the time--layer grid raises mIoU from 31.78 to 33.32.
Adding a heuristic top-$b$ subset reaches 32.82, while the unnormalized joint
policy reaches 31.68 (Figure~\ref{fig:action-ablation}). Expanding the action
space alone therefore does not improve accuracy in this setting.

\textbf{Channel-score normalization.}
With $A_l$ fixed, \Policy{} reaches $37.7\pm0.7$ mIoU, compared with
$36.8\pm0.6$ without the $1/b$ channel-score factor
(Table~\ref{tab:fixed-alignment}). Both use the same three seeds and training
settings. This comparison is separate from the single-seed action-axis
ablation and the main segmentation table.

\begin{figure}[t]
\centering
\includegraphics[width=\linewidth]{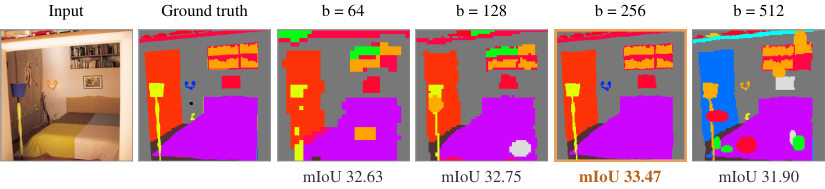}
\vspace{-1.5em}
\caption{Effect of the channel budget on ADE20K segmentation. Channels are selected
by activation magnitude at a fixed timestep and layer. A separate student is
trained for each budget. Predictions are shown for one validation image, while
mIoU is computed over the validation set. No learned policy is used.}
\label{fig:budget-visual}
\end{figure}

\textbf{Channel budget.}
The sweep in Figure~\ref{fig:budget-visual} uses top-magnitude selection at a
fixed timestep and layer, with a separate student trained for each budget.
It does not use a learned policy. Increasing $b$ from 128 to 256 raises the
best validation mIoU by 0.72 and the mean over the last five epochs by 0.27.
The sweep changes both the selected features and the KD-loss scale, so it
does not identify an optimal budget independently of distillation strength
(Appendix~\ref{app:budget}). We retain $b=128$ for the compact representation
used in the selector comparisons.

\subsection{Transformer teachers}
\label{sec:transfer}

We transfer the action to teachers with an ordered feature lattice. A U-Net
teacher \citep{ronneberger2015} uses denoising steps, layers, and channels. For
a DiT, MMDiT, or flow transformer \citep{peebles2023,esser2024,lipman2023}, let
$H^{\rm img}_{\lambda,d}(x)\in\R^{N\times D}$ denote the image-token state at
log-SNR or flow coordinate $\lambda$ and depth $d$. The corresponding selected
target is:
\begin{equation}
\widetilde G_{\lambda,d,K}(x)=
\mathcal R_d\!\left(A_d\!\left(\operatorname{LN}
\left(H^{\rm img}_{\lambda,d}(x)\right)_{:,K}\right)\right),
\qquad K\subseteq[D],\quad |K|=b,
\label{eq:transformer-action}
\end{equation}
where $A_d$ maps retained hidden dimensions to the common target width and
$\mathcal R_d$ restores the spatial grid. In MMDiT, the action applies only to
the image stream after joint attention. The student loss is retained, and the
subset still has exactly $b$ coordinates. Model cards are linked for
\href{https://huggingface.co/runwayml/stable-diffusion-v1-5}{SD-1.5},
\href{https://huggingface.co/PixArt-alpha/PixArt-Sigma-XL-2-1024-MS}{PixArt-$\Sigma$},
and \href{https://huggingface.co/stabilityai/stable-diffusion-3.5-medium}{SD3.5
Medium}.

\begin{table}[t]
\caption{Separate matched-fraction interface evaluation across teacher
architectures. The selected coordinate changes from U-Net channels to
image-token hidden dimensions. These rows are not a continuation of
Table~\ref{tab:segmentation}: each uses its teacher-specific feature interface.}
\vspace{0.5em}
\label{tab:transformer-results}
\centering
\begin{tabular*}{\linewidth}{@{\extracolsep{\fill}}lccccc@{}}
\toprule
Teacher & Selected coordinates & $b/D$ & ADE20K & CIFAR-100 & SPair-71k \\
\midrule
SD-1.5 U-Net & $(t,l,K_{\rm ch})$ & $128/1280$ & $34.1$ & $70.8$ & $72.8$ \\
PixArt-$\Sigma$ DiT & $(\lambda,d,K_{\rm hid})$ & $115/1152$ & $35.0$ & $71.6$ & $73.5$ \\
SD3.5 MMDiT & $(\lambda,d,K_{\rm img})$ & $154/1536$ & $36.4$ & $72.9$ & $74.6$ \\
\bottomrule
\end{tabular*}
\end{table}

We instantiate the transformer interfaces with PixArt-$\Sigma$
\citep{pixart2024} and SD3.5 MMDiT \citep{esser2024}. The matched fractions
in Table~\ref{tab:transformer-results} lie between
$9.98\%$ and $10.03\%$ of the available feature width. This keeps the hard
selection pressure comparable when a U-Net channel axis is replaced by a
transformer hidden-dimension axis. The table is therefore evidence for the
interface under a matched capacity fraction, not a ranking based on raw model
width. It is reported separately from Tables~\ref{tab:segmentation}
and~\ref{tab:recognition}. No numerical contrast is drawn across these
teacher-specific evaluations.

\FloatBarrier

\subsection{Limitations}

\textbf{Channel budget.} The budget sweep in Figure~\ref{fig:budget-visual}
changes the selected features and the distillation strength together. The
masked KD loss is averaged over all 1280 channels, so its mean over epochs
10--19 grows from 0.59 at $b=128$ to 0.94 at $b=256$ and 1.35 at $b=512$. The
drop at $b=512$ therefore cannot be attributed to redundant channels.
A follow-up sweep with KD normalized over the selected channels would control
this change in loss scale
(Appendix~\ref{app:budget}).

\textbf{Repetitions.} The normalization ablation and the RepFusion
comparison are repeated over three seeds (Table~\ref{tab:fixed-alignment}).
The action-axis ablation, the budget sweep, and the qualitative analysis use a
single seed.

\textbf{Transformer teachers.} Table~\ref{tab:transformer-results} matches
the capacity fraction across teachers but uses one evaluation per teacher. It
supports the transferred action interface, not a ranking of teacher
architectures.

\section{Conclusion}
\label{sec:conclusion}

We presented \Policy{}, which treats diffusion feature extraction as a
budgeted decision: a policy jointly selects the timestep, the layer, and an
exact-size channel subset of a frozen teacher, and the selected features
distill a student that runs alone at inference. The experiments support three
conclusions. First, joint selection pays off at a fixed budget: with 128 of
1280 channels, \Policy{} exceeds full-width extraction at a fixed
extraction point on all four segmentation benchmarks and exceeds RepFusion on
CIFAR-10 and CIFAR-100. Second, channel-score normalization improves the
tested joint policy. Without the $1/b$ factor, the joint policy falls below the
timestep-only policy in the action-axis ablation. With fixed alignment,
normalization raises
ADE20K mIoU from 36.8 to 37.7 over three seeds. Third, the action transfers
to DiT and MMDiT teachers by replacing U-Net channels with image-token hidden
dimensions while retaining the student loss and exact-cardinality constraint. Separating
channel count from distillation strength remains the main open question for
choosing the operating budget.

\bibliographystyle{icomp2026_conference}
\bibliography{policy_mask}

\clearpage
\appendix
\section{Experimental details}
\label{app:details}

\subsection{Channel-budget sweep}
\label{app:budget}
Figure~\ref{fig:budget-visual} compares channel budgets on ADE20K using
top-magnitude selection at a fixed timestep and layer, with a separate student
trained for each budget. The highest validation mIoU is obtained at $b=256$.
The masked distillation loss is averaged over all 1280 channels, so changing
the budget also changes its scale. This experiment does not isolate the effect
of channel count from the change in distillation strength.

\subsection{Fixed alignment}
\label{app:alignment}
With trainable $A_l$, the KD loss fell below 0.005 within two epochs in each
diagnostic run. This behavior is consistent with projection shrinkage toward
zero. We therefore fixed $A_l$ for the channel-score normalization ablation
and the RepFusion comparison.

\section{Additional results}

\begin{table}[h]
\begin{minipage}[b]{0.48\linewidth}
\caption{Fixed-slot reference runs on ADE20K. Each method is trained for
14 epochs with one seed. The reported metric is the best validation mIoU.}
\label{tab:fixed-references}
\vspace{0.5em}
\begin{tabular*}{\linewidth}{@{\extracolsep{\fill}}lrr@{}}
\toprule
Selector & Width & mIoU \\
\midrule
All channels & 1280 & 38.35 \\
Random subset & 128 & 37.88 \\
Top-variance subset & 128 & 37.70 \\
\bottomrule
\end{tabular*}
\end{minipage}\hfill
\begin{minipage}[b]{0.48\linewidth}
\setlength{\tabcolsep}{3pt}
\caption{ADE20K results with fixed $A_l$ after 14 training epochs.
Values are the mean and sample standard deviation of the best validation
mIoU over three seeds.}
\label{tab:fixed-alignment}
\vspace{0.5em}
\begin{tabular*}{\linewidth}{@{\extracolsep{\fill}}lrr@{}}
\toprule
Method & Width & mIoU \\
\midrule
\Policy{} & 128 & $37.7\pm0.7$ \\
Unnormalized joint policy & 128 & $36.8\pm0.6$ \\
RepFusion & 1280 & $33.1\pm0.3$ \\
\bottomrule
\end{tabular*}
\end{minipage}
\end{table}

\subsection{Fixed-slot reference runs}
\label{app:references}
Table~\ref{tab:fixed-references} compares feature selection methods on ADE20K
at a fixed timestep and layer. The full-width baseline uses 1280 channels,
while random and top-variance selection retain 128 channels.
These 14-epoch reference runs follow a separate protocol and are not compared
numerically with Table~\ref{tab:segmentation}.

\subsection{Channel-score normalization and RepFusion comparison}
\label{app:normalization}
The normalized and unnormalized policies use fixed $A_l$, the same seeds,
and the same training settings. They differ only in the $1/b$ channel-score
factor. Our ADE20K implementation of RepFusion uses SD-1.5 and follows two
stages: feature distillation followed by segmentation training. The
teacher-side alignment remains fixed. RepFusion uses all 1280 dimensions,
while \Policy{} retains 128.

\section{Qualitative and class-wise analysis}
\label{app:qualitative}
\begin{figure}[htbp]
\centering
\includegraphics{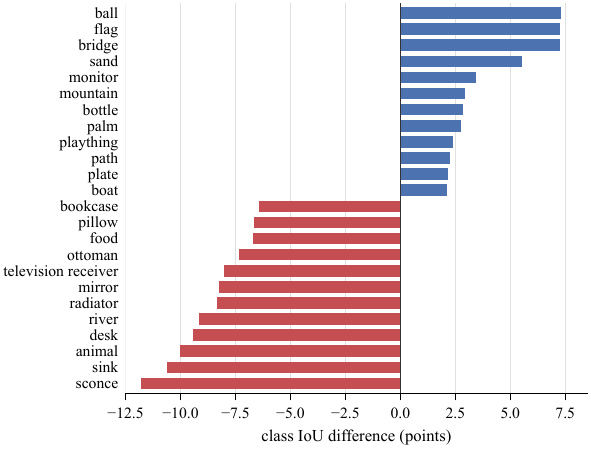}
\caption{Class IoU differences between \Policy{} and RepFusion on ADE20K,
seed 0. The plot shows the 12 largest gains and losses among classes present
in at least 20 validation images. Positive values favor \Policy{}.}
\label{fig:class-analysis}
\end{figure}

The seed-0 comparison covers 2000 ADE20K validation images. \Policy{}
has higher image-mIoU than RepFusion on 39.1\% of images. This image-level
comparison is separate from the dataset mIoU reported across three seeds.
Class IoU improves for ball, flag, bridge, sand, monitor, and mountain,
with losses for sconce, sink, animal, and desk (Figure~\ref{fig:class-analysis}).
The gain and loss panels show the 12 largest differences in each direction.

\begin{figure}[htbp]
\centering
\includegraphics[width=\linewidth]{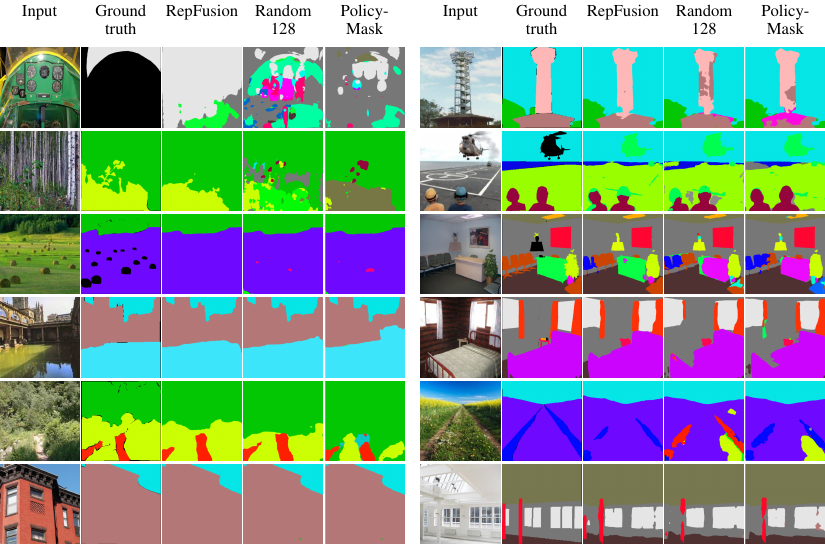}
\caption{ADE20K validation, seed 0. The 12 images with the largest losses
in image-mIoU relative to RepFusion. Selected by outcome. Columns follow
Figure~\ref{fig:qualitative}.}
\label{fig:panel-loss}
\end{figure}

\end{document}